\documentclass[fleqn,10pt]{wlscirep}
\usepackage[utf8]{inputenc}
\usepackage[T1]{fontenc}
\usepackage{lineno}
\usepackage{amsmath} 
\usepackage{multirow}
\usepackage{amsmath}
\usepackage{graphicx} 

\title{A Digital Pathology Resource for Liver Cancer Quantification with Datasets, Benchmarks, and Tools}

\author[1,$\dag$]{Ying Xiao}
\author[2,$\dag$]{Shimiao Tang}
\author[2,$\dag$]{Xitong Ling}
\author[2,]{Weiming Chen}
\author[2,]{Jun Wang}
\author[2,]{Jiawen Li}
\author[2,]{Huaitian Yuan}
\author[1,]{Jianghui Yang}
\author[1,]{Bowen Li}
\author[1,]{Huan Li}
\author[1,]{Yiting Meng}
\author[2]{Tian Guan}
\author[2,*]{Yonghong He}
\author[1,*]{Hongfang Yin}
\affil[1]{Department of Pathology, Beijing Tsinghua Changgung Hospital, School
of Clinical Medicine, Tsinghua Medicine, Tsinghua University, Beijing, China}
\affil[2]{Shenzhen International Graduate School, Tsinghua University, Shenzhen, China}

\affil[*]{corresponding author(s): Hongfang Yin (yhfa00530@btch.edu.cn), Yonghong He (heyh@sz.tsinghua.edu.cn) }

\affil[$\dag$]{these authors contributed equally to this work}

\begin{abstract}
Liver cancer, especially hepatocellular carcinoma (HCC), imposes a substantial global disease burden. Accurate diagnosis and prognostic assessment directly influence treatment selection and patient survival, and pathological examination remains the gold standard for liver cancer diagnosis. Identifying diverse tissue components and pathological subtypes on histopathology slides is crucial for estimating postoperative recurrence risk and overall prognosis. However, most publicly available resources are still provided at the whole-slide image (WSI) level, and well-annotated datasets for fine-grained tissue component identification in liver cancer are scarce, which hinders reproducible model development and the deployment of quantitative analysis tools. To address this gap, we release HepatoBench, a patch-level image database for liver cancer with annotations for seven key tissue categories. Based on HepatoBench, we train and open-source a deep learning classification model as a tissue recognition tool. Furthermore, we train a WSI-level tumor/non-tumor segmentation model to automatically localize lesion regions across entire slides. By integrating the patch-level tissue classifier with the WSI-level segmentation model, we build HepatoQuant, an end-to-end, disease-specific regional quantification tool for liver cancer, enabling a unified workflow from WSIs to tissue composition parsing and quantitative statistics. We also open-source HepatoBench, the benchmarking protocol, and supporting tools, providing a solid foundation for automated regional quantification and fair method comparison in liver cancer pathology.

\end{abstract}
\begin{document}

\flushbottom
\maketitle
\thispagestyle{empty}   
\section*{Background \& Summary}

Primary liver cancer, especially hepatocellular carcinoma (HCC), is one of the most common primary liver malignancies worldwide~\cite{ferlay2015cancer}, with persistently high incidence and mortality rates. According to GLOBOCAN and recent epidemiological reviews, liver cancer ranks among the leading causes of cancer and cancer-related deaths globally, with significant variations in incidence and mortality across different countries/regions.  It is projected that the overall number of cases will continue to rise in the coming decades due to population growth and aging~\cite{bray2024global}. Clinically, liver cancer often presents with subtle symptoms in its early stages, and is frequently diagnosed at an advanced stage, leading to a generally poor prognosis. Although significant progress has been made in surgical resection, local ablation, targeted therapy, and immunotherapy for liver cancer in recent years, liver cancer typically develops against a background of chronic liver disease, exhibiting distinct multi-stage and multi-factorial characteristics~\cite{kanda2019molecular, kovalic2019nonalcoholic, oh2023latest, lampimukhi2023review, rayapati2025environmental}, and showing significant variations in treatment response among individual patients. How to more accurately diagnose, assess prognosis, and guide individualized treatment remains one of the core challenges in the field of liver cancer diagnosis and treatment.Pathological examination is widely recognized as the "gold standard" for the diagnosis and classification of liver cancer.  The precise quantification of different tissue components in pathological sections is closely related to the biological behavior and clinical prognosis of liver cancer~\cite{tang2001hepatocellular, edmondson1954primary, balogh2016hepatocellular, sas2022tumor, kuo2023tumor, libbrecht2002hepatic, salomao2010steatohepatitic}. However, traditional liver cancer pathological assessment is primarily based on manual experience, relying mainly on qualitative or semi-quantitative judgments of whole sections or local areas by pathologists. This approach suffers from limitations such as strong subjectivity, limited reproducibility, and difficulty in fine-grained quantification in a clinical context. Objective and reproducible large-scale quantitative analysis of these tissue components based on pathological models can help to more precisely characterize the histological heterogeneity of liver cancer.  Furthermore, it can reveal potential new pathological quantitative indicators and provide reliable basic features for multimodal data fusion. In summary, developing automated and standardized quantitative analysis methods for liver cancer pathology has significant clinical and research value.

The development of large-scale models has greatly facilitated digital pathology research~\cite{berbis2023computational, bankhead2022developing}. By pre-training on a large number of pathological images, models can support a wide range of downstream applications, including classification, segmentation, and weakly supervised learning, demonstrating strong generalization ability across different tissue types and diagnostic tasks. Recent studies ~\cite{liang2023deep, lou2025large, ling2025comprehensive}have validated the feasibility of basic pathology models: models trained on massive histopathological image databases can exhibit excellent performance in various clinical tasks.

The reproducibility and high performance of models require high-quality, large-scale publicly available pathology datasets~\cite{he2022masked}. Compared to the natural image domain, medical pathology data is more expensive to acquire and more difficult to annotate, and there are significant differences between different centers in staining procedures, scanning equipment, and diagnostic criteria. Therefore, standardized, shareable publicly available pathology datasets play an irreplaceable role in method evaluation, model comparison, and cross-institutional generalization studies. Currently, some international public databases (such as TCGA, etc~\cite{cooper2018pancancer, kim2021paip, bejnordi2017diagnostic, 8447230}) provide digital pathology slides, including those for liver cancer, along with corresponding clinical and molecular information~\cite{weinstein2013cancer}, providing important resources for large-scale liver cancer pathology analysis. However, high-quality annotated data for fine-grained quantification of liver cancer tissue remains relatively scarce, and the lack of a unified benchmark among different studies limits the systematic comparison and practical translation of related methods. Systematically integrating publicly available data resources, constructing standardized datasets, and developing evaluation tools for quantitative analysis of liver cancer pathology are of great significance for promoting the in-depth application of basic pathology models in the field of liver cancer~\cite{asif2023unleashing}.

In this paper, we filtered and removed duplicate, ambiguous, and poorly stained patches from our self-constructed dataset. Using this dataset, we re-evaluated the quantitative performance using three pathology-specific pre-trained feature encoders: UNI-v1\cite{chen2024towards}, Virchow-V2\cite{vorontsov2024foundation} and Gigapath\cite{xu2024whole}. We further annotated and open-sourced a patch-level liver tissue dataset with seven categories, including Tumor (TUM), Fibrosis (FIB), Inflammation (INF), Necrosis (NEC), Normal tissue (NOR), Bile duct reaction (REA), and Steatosis (STE). Based on this dataset, we trained linear-probe models on top of these four encoders and benchmarked their performance to compare the representational capacity of different foundation models for liver tissue recognition. To quantify tissue-type composition within tumor and non-tumor regions at the whole-slide level, we additionally trained a WSI-level region segmentation model via LoRA fine-tuning ~\cite{bolhassani2025lora}on top of a pathology foundation model. Finally, by integrating the patch-level tissue classifier and the WSI-level segmentation model, we developed an end-to-end liver cancer tissue quantification and scoring system, and we release the corresponding model weights and inference code.

\section*{Technical Validation}

\subsection*{Dataset Overview}
We release a patch-level liver histopathology dataset curated for fine-grained tissue composition analysis in liver cancer.
All patches are extracted at 20$\times$ magnification with a fixed spatial resolution of $150 \times 150$ pixels.
Each patch is assigned to one of seven tissue categories: Tumor (TUM), Fibrosis (FIB), Inflammation (INF),
Necrosis (NEC), Normal tissue (NOR), Bile duct reaction (REA), and Steatosis (STE).
To improve label reliability, we performed quality control to remove duplicate patches, ambiguous regions, and poorly stained samples
(e.g., strong blur, severe staining artifacts, or out-of-focus areas), yielding a clean benchmark set for quantitative evaluation.
The dataset contains 140{,}000 annotated patches in total (example numbers), with the following class distribution:
TUM 13{,}200, FIB 26{,}364, INF 4{,}776, NEC 22{,}023, NOR 13{,}200, REA 1{,}540, and STE 6{,}198.
We split the dataset into training/validation/testing subsets with a ratio of 7:2:1 for pathology foundation models evaluation.

\subsection*{Exclusion Criteria}
To ensure data quality, label reliability, and fairness in benchmark testing, we applied predefined exclusion criteria at the slide and patch levels before dataset release and model evaluation. Specifically, we excluded the following situations: (i) patches identified as duplicates or near-duplicates through comparing file MD5 values and fine-grained determination within the same Hash bucket; (ii) low-quality patches affected by scanning or preparation artifacts, including severe out-of-focus/blurring, severe staining failure or extreme color deviation, tissue wrinkles, tears, bubbles, handwriting, severe compression artifacts, large fragments, or significant background/blank areas exceeding a preset threshold; (iii) patches containing only edge tissue fragments or edge areas
In addition, we ensured that the train/validation/test partitions are performed at the slide level.These exclusion rules produce a clean, standardized, and reproducible benchmark suitable for quantitative evaluation of liver tissue composition in digital pathology.

\section*{Data Records}

The HepatoBench dataset is a seven-class histopathology image benchmark curated for evaluating pathology foundation models. It comprises images from seven tissue categories: TUM (13,200), FIB (26,364), INF (4,776), NEC (22,023), NOR (13,200), REA (1,540), and STE (6,198). We split the dataset into training, validation, and testing subsets at a ratio of 7:2:1 for standardized model evaluation. The HepatoBench dataset can be accessed at HuggingFace (\hyperlink{https://doi.org/10.57967/hf/8231}{https://doi.org/10.57967/hf/8231})~\cite{ying_xiao_2026}. We publicly release the raw images, class labels, and the official dataset partition. Images are organized into class-specific directories, where each folder name corresponds to its category label. The dataset meta information, including the basic attributes, structure, and relevant metadata of each dataset, is provided in \texttt{dataset\_meta.json}.

\section*{Methods}

\subsection*{Methodology}
\begin{figure}[htbp]
\centering
\includegraphics[width=0.94\linewidth]{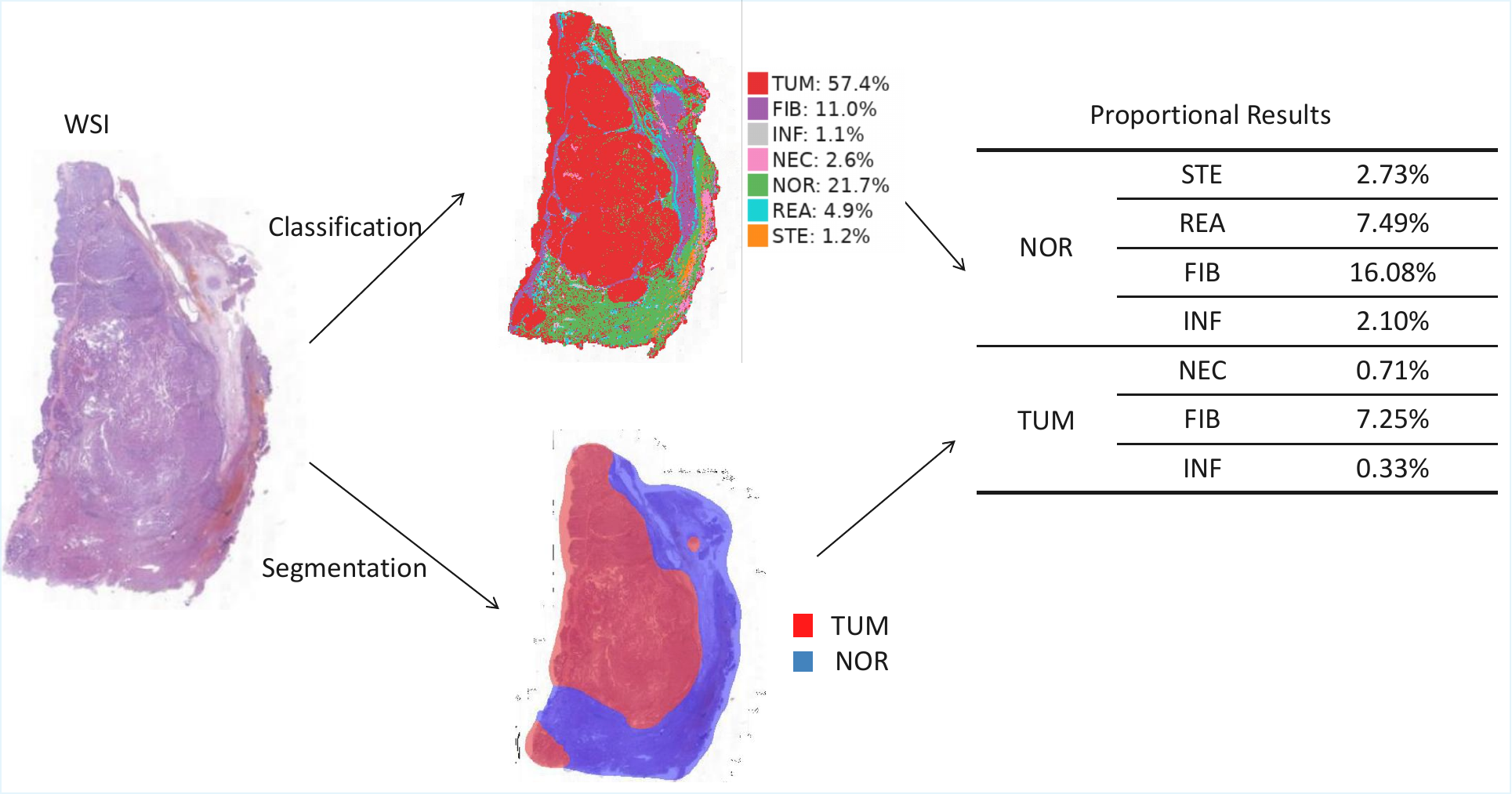}
\caption{End-to-end quantitative analysis pipeline for hepatocellular carcinoma whole slide images.}
\label{end2end}
\end{figure}

As shown in Figure \ref{end2end}, We build an end-to-end pipeline for quantitative liver cancer pathology analysis with three components: patch-level liver tissue region classification, whole-slide region segmentation, and slide-level quantitative tissue composition analysis. For each whole slide image (WSI), we first extract tissue patches through a standard tiling procedure and then classify the tissue patches. Patch classification is achieved by using a frozen pathology foundation model (PFM) as a feature extractor and training a lightweight classification head on top of the extracted embeddings to predict seven liver tissue categories. In parallel,  we utilize a segmentation model based on PFM, fine-tune it using LoRA and frozen backbone to achieve regional segmentation of the entire slide, obtaining a mask of tumor/non-tumor regions for the entire slide. Finally, we integrate the prediction results of patch-level tissue categories and the regional mask of the entire slide to calculate the statistical data of tissue category proportions within the tumor/non-tumor regions of the entire slide, thereby achieving standardization and repeatable quantification within tumor and non-tumor regions
\subsection*{Data Preprocessing}
To construct the patch-level dataset, board-certified pathologists cropped multiple non-overlapping image patches (150 × 150 pixels) at 20× magnification from selected regions of WSIs and assigned a tissue category to each patch. This resulted in a self-curated liver histopathology dataset comprising seven tissue classes: Tumor (TUM), Fibrosis (FIB), Inflammation (INF), Necrosis (NEC), Normal Tissue (NOR), Biliary Reaction (REA), and Steroidal Degeneration (STE). Prior to model training, we performed dataset curation to remove low-quality and redundant samples. Specifically, we eliminated patches with ambiguous morphology, and patches with poor staining or severe artifacts (e.g., blur, tissue folds/wrinkles, pen markings, and scanning noise), then we applied MD5-based duplicate detection, followed by fine-grained screening within each hash bucket to identify and exclude duplicates or near-duplicates patches. This cleaning procedure reduced redundancy and improved the reliability of the benchmark.

For whole-slide processing, pathologists additionally performed semantic segmentation on 111 WSIs to delineate tumor and normal regions and to obtain corresponding contour coordinates. Before training, WSIs were read at a fixed magnification and conducted foreground segmentation to exclude background regions, we then generated mask maps by integrating the annotated contour coordinates with the foreground tissue regions, thereby producing tumor/non-tumor regions overlaid on the original WSI. Based on processed datasets, we employed three pretrained feature encoders—UNI, GigaPath, and Virchow-V2 to extract pixel-level representations, which were used to support seven-class tissue label prediction for WSI-derived patches. 

During training, we applied standard histopathology augmentations, including random resizing/cropping, horizontal/vertical flipping, rotation, and mild color jitter, to improve robustness to staining and scanner variability. Data splits were performed at the slide/patient level to prevent information leakage, ensuring that patches from the same WSI did not appear in different splits. Moreover, WSIs used to derive patches for the seven-class classification dataset were excluded from the test set. This design avoids confounding when evaluating the end-to-end inference pipeline that combines segmentation and classification to estimate tissue composition within tumor and normal regions, including the proportions of STE, REA, FIB, and INF in normal regions and NEC, FIB, and INF in tumor regions.

\subsection*{Liver Tissue Region Classification}
\begin{figure}[htbp]
\centering
\includegraphics[width=0.94\linewidth]{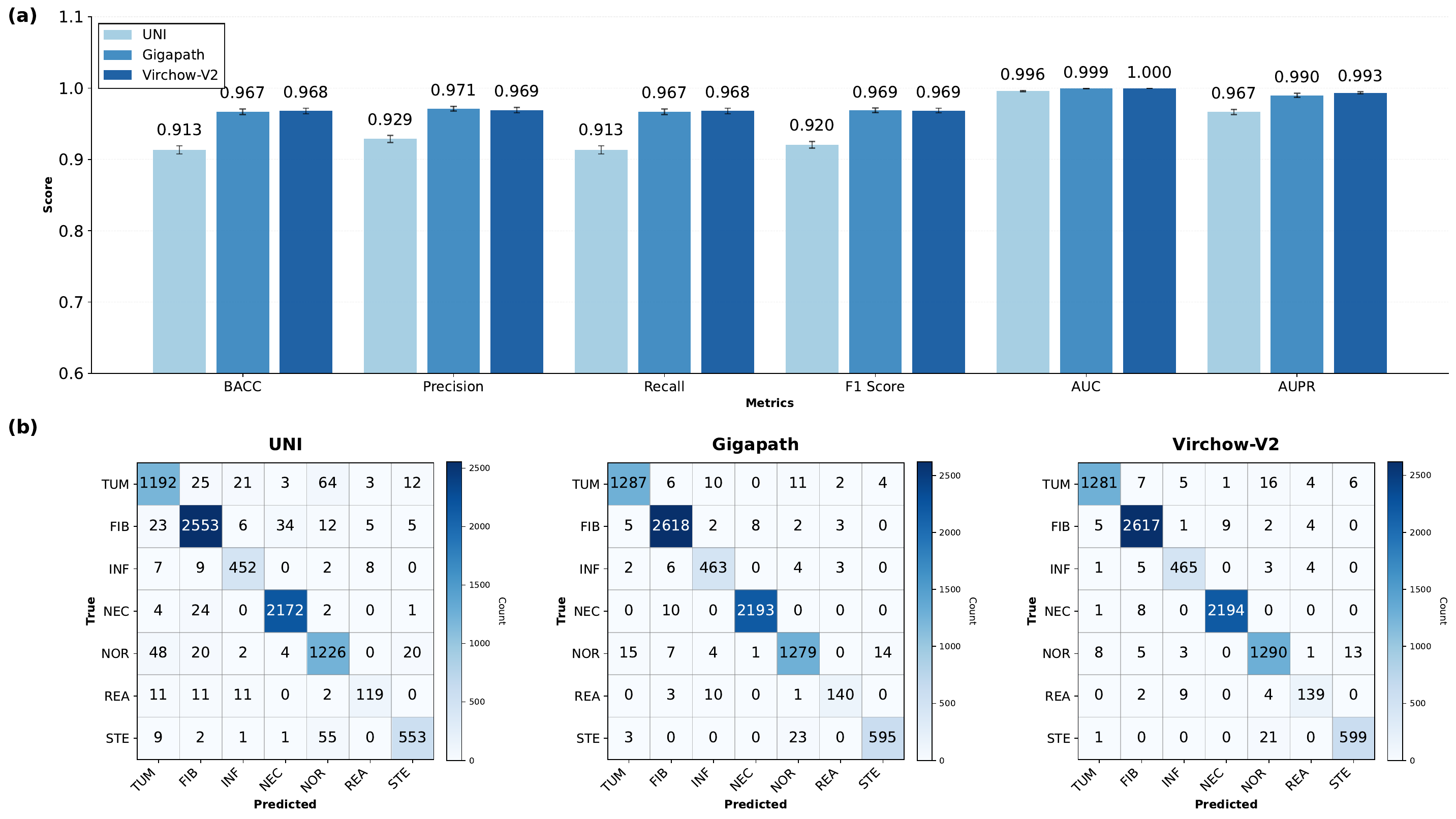}
\caption{Benchmark results of pathology foundation models on HepatoBench.}
\label{performance}
\end{figure}

As shown in Figure \ref{performance}, We benchmark multiple state-of-the-art pathology-specific PFMs (UNI, Virchow-V2, Gigapath) under a unified linear-probe setting to compare their representation quality for liver tissue recognition. For each encoder, we freeze all backbone parameters and use the model only to extract patch embeddings. On top of these frozen features, we train a lightweight classification head (a single linear layer) to predict the seven tissue categories. With this design, only the classification head is updated during training, which makes training efficient and ensures that performance differences mainly reflect the intrinsic representational capacity of the pretrained encoder rather than task-specific fine-tuning of the backbone.

At inference time, a WSI is tiled into tissue patches, each patch is encoded by the frozen PFM, and the trained head outputs a tissue-type prediction. The patch-level predictions can be aggregated into a dense tissue map over the whole slide, providing fine-grained morphological composition cues for subsequent quantific

\subsection*{Whole Slide Region Segmentation}
To quantify tissue composition within clinically meaningful regions, we additionally train a whole-slide region segmentation model to delineate tumor and non-tumor areas (and optionally exclude background). The segmentation backbone is initialized from a pathology foundation model to leverage strong pretrained representations. Instead of full fine-tuning, we adopt LoRA to adapt the foundation model with a small number of trainable parameters, and train the LoRA modules together with a lightweight segmentation head/decoder. This strategy reduces training cost and memory usage while preserving the generalization ability of the pretrained backbone.

Because WSIs are extremely large, both training and inference of patch-level seven classification are conducted in a tile-based sliding-window manner. Predicted tile masks are stitched back to the WSI coordinate space to produce a complete whole-slide region mask. Overlapping windows can be used to reduce boundary artifacts and improve mask continuity.

\subsection*{Quantitative Analysis of Liver Tissue}
We perform quantitative tissue analysis by integrating the patch-level seven-class tissue predictions with the WSI-level tumor/non-tumor segmentation mask. Specifically, the segmentation model provides a region partition of the slide, while the patch classifier provides fine-grained tissue labels for each patch. By mapping each patch prediction back to its spatial position on the slide and referencing the segmentation mask, we assign each patch to either the tumor region or the non-tumor region.

Based on this region-aware assignment, we summarize tissue composition separately within tumor and non-tumor areas to obtain standardized quantitative descriptors of liver histology, such as the relative presence of fibrosis, inflammation, necrosis, steatosis, bile duct reaction, and normal tissue. The final system outputs the whole-slide region mask, the patch-level tissue map, and a structured tissue composition report, assisting objective, reproducible, and scalable quantification for liver cancer pathology studies and downstream multimodal analyses.

\subsection*{Evaluation metrics}
For the patch-level seven-class classification task, we evaluate model performance using Accuracy (Acc), Precision, Recall and F1-score.
For whole-slide image (WSI) region segmentation, we report Intersection-over-Union (IoU) ,Dice coefficient (Dice), Precision, Recall and F1-score as the primary evaluation metrics.

\section*{Usage Notes}

The HepatoBench dataset is publicly available under the Creative Commons Attribution-NonCommercial 4.0 (CC BY-NC 4.0) license. However, please note that this dataset is not intended for developing diagnosis-focused algorithms or models, and it should not be used as the sole basis for clinical evaluation in classification tasks.

\section*{Code availability}

We perform whole-slide image (WSI) segmentation by LoRA-based fine-tuning of pathology foundation models, and publicly release the corresponding codebase at
\href{https://github.com/lingxitong/PFM_Segmentation}{https://github.com/lingxitong/PFM\_Segmentation}.

\bibliography{sample}

\section*{Acknowledgements} 

This work was supported by the Beijing Research Ward Excellence Program (BRWEP2024W032240114), the National Natural Science Foundation of China (NSFC) under Grant No. 82430062.

\section*{Author contributions statement}

Y.X., S.T., and X.L. contributed equally to this work. Y.X., S.T., X.L., and T.G. conceived the study and designed the experiments. Y.X., S.T., and X.L. performed the experiments and analyzed the results. W.C., J.W., J.L., H.Y., J.Y., B.L., H.L., and Y.M. contributed to dataset correction and construction. Y.H., and H.Y. supervised the study, contributed to manuscript preparation, and provided critical insights into the manuscript organization and presentation. All authors reviewed and approved the final manuscript.

\section*{Competing interests} 
The authors declare no competing interests.
\newpage
\section*{Figures \& Tables}




\end{document}